\documentclass[runningheads]{llncs}
\usepackage{graphicx}

\usepackage{hyperref}
\usepackage{booktabs}
\usepackage{multirow}
\usepackage{array}
\newcommand{\tabincell}[2]{\begin{tabular}{@{}#1@{}}#2\end{tabular}}
\usepackage{amsmath,amssymb,amsfonts} 
\usepackage[table]{xcolor}

\mathchardef\mhyphen="2D

\usepackage{tikz}
\usepackage{comment}
\usepackage{amsmath,amssymb} 
\usepackage{color}
\usepackage[misc]{ifsym}

\usepackage[ruled,vlined,linesnumbered]{algorithm2e}

\begin{document}
\pagestyle{headings}
\mainmatter
\def\ECCVSubNumber{655}  

\title{Lidar Point Cloud Guided Monocular 3D Object Detection} 

\titlerunning{Lidar Point Cloud Guided Monocular 3D Object Detection}
\authorrunning{L. Peng et al.}
%


\author{Liang Peng\inst{1,2}, Fei Liu\inst{3}, Zhengxu Yu\inst{1}, Senbo Yan\inst{1,2}, Dan Deng\inst{2},\\ Zheng Yang\inst{2},  Haifeng Liu\inst{1}, and Deng Cai\inst{1,2} \textsuperscript{\Letter}
}
\institute{
State Key Lab of CAD\&CG, Zhejiang University, China  \\
\email{\{pengliang, senboyan, haifengliu\}@zju.edu.cn yuzxfred@gmail.com dengcai@cad.zju.edu.cn} \\
\and Fabu Inc., Hangzhou, China\\
\email{\{dengdan, yangzheng\}@fabu.ai} 
\and State Key Lab of Industrial Control and Technology, Zhejiang University, China  \\
\email{liufei21@zju.edu.cn}
}

\maketitle

\begin{abstract}
	Monocular 3D object detection is a challenging task in the self-driving and computer vision community.
	As a common practice, most previous works use manually annotated 3D box labels, where the annotating process is expensive.
	In this paper, we find that the precisely and carefully annotated labels may be unnecessary in monocular 3D detection, which is an interesting and counterintuitive finding.
	Using rough labels that are randomly disturbed, the detector can achieve very close accuracy compared to the one using the ground-truth labels.
	We delve into this underlying mechanism and then empirically find that: concerning the label accuracy, the 3D location part in the label is preferred compared to other parts of labels.
	Motivated by the conclusions above and considering the precise LiDAR 3D measurement, we propose a simple and effective framework, dubbed LiDAR point cloud guided monocular 3D object detection (LPCG).
	This framework is capable of either reducing the annotation costs or considerably boosting the detection accuracy without introducing extra annotation costs.
	Specifically, It generates pseudo labels from unlabeled LiDAR point clouds.
	Thanks to accurate LiDAR 3D measurements in 3D space, such pseudo labels can replace manually annotated labels in the training of monocular 3D detectors, since their 3D location information is precise.
	LPCG can be applied into any monocular 3D detector to fully use massive unlabeled data in a self-driving system.
	As a result, in KITTI benchmark, we take the first place on both monocular 3D and BEV (bird's-eye-view) detection with a significant margin.
	In Waymo benchmark, our method using 10\% labeled data achieves comparable accuracy to the baseline detector using 100\% labeled data. 
	The codes are released at \href{https://github.com/SPengLiang/LPCG}{https://github.com/SPengLiang/LPCG}.
\end{abstract}

\section{Introduction}
	3D object detection plays a critical role in many applications, such as self-driving.
	It gives cars the ability to perceive the world in 3D, avoiding collisions with other objects on the road. 
	Currently, the LiDAR (Light Detection and Ranging) device is typically employed to achieve this \cite{PART,PP,PVRCNN,se-ssd}, with the main shortcomings of the high price and limited working ranges. 
	The single camera, as an alternative, is widely available and several orders of magnitude cheaper, consequently making monocular methods \cite{Mono_PseudoLidar,M3D,D4LCN,MonoFlex} popular in both industry and academia.
	
\begin{figure}[h]
		\begin{center}
				\includegraphics[width=0.75\linewidth]{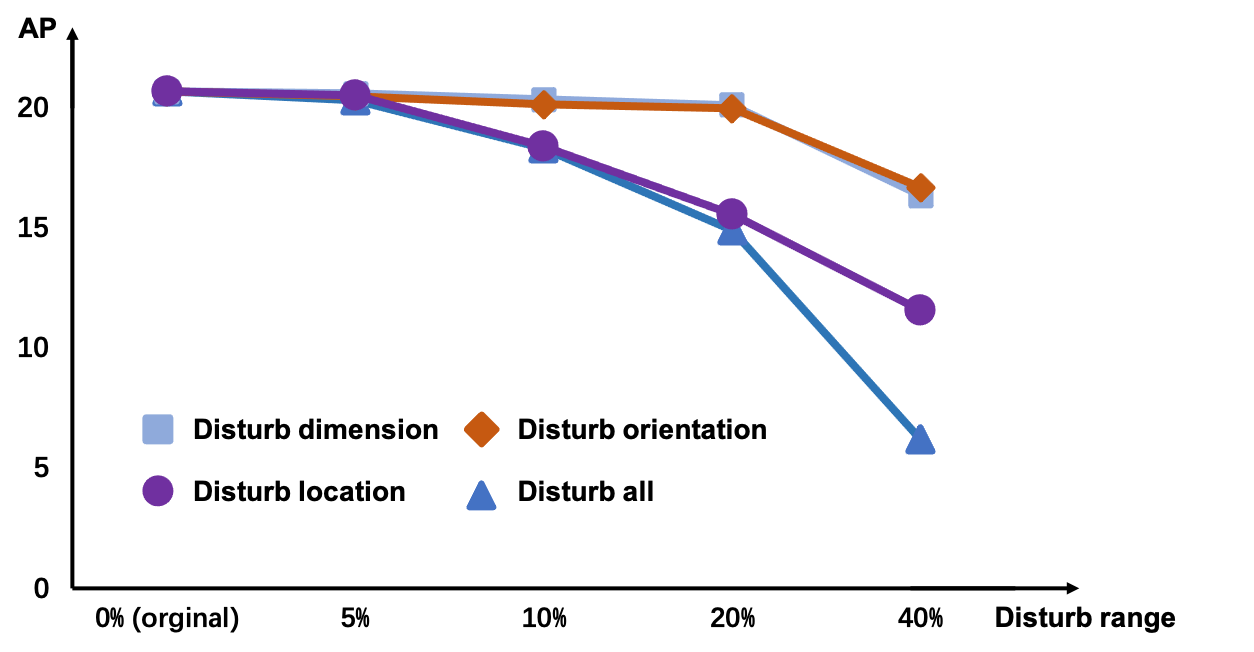}
		\end{center}
		\caption{
		We disturb the perfect manually annotated labels by randomly shifting the corresponding values within the percentage range.
		We can see that: 1) the disturbed labels (5\%) and perfect labels  lead to close accuracy; 2) the location dominates the overall accuracy (10\%, 20\%, 40\%).
		} 
		\label{fig: dis}
	\end{figure}

	To the best of our knowledge, most previous monocular-based works  \cite{M3D,D4LCN,CaDDN,MonoFlex} employ the precisely annotated 3D box labels.
	The annotation process operated on LiDAR point clouds is time-consuming and costly.
	In this paper, we empirically find that {\bf the perfect manually annotated 3D box labels are not essential in monocular 3D detection}.
	We disturb the manually annotated labels by randomly shifting their values in a range, while the detector respectively trained by disturbed labels and perfect labels show very close performance.
	This is a counterintuitive finding.
	To explore the underlying mechanism, we divide a 3D box label into different groups according to its physical nature (including 3D locations, orientations, and dimensions of objects), and disturb each group of labels, respectively.
	We illustrate the experiment in Figure \ref{fig: dis}.
	The results indicate that the precise location label plays the most important role and dominates the performance of monocular 3D detection, and the accuracy of other groups of labels is not as important as generally considered.
	The underlying reason lies in the ill-posed nature of monocular imagery.
	It brings difficulties in recovering the 3D location, which is the bottleneck for the performance.

	Unlike other classical computer vision tasks, manually annotating 3D boxes from monocular imagery is infeasible. 
	It is because the depth information is lost during the camera projection process.
	Actually, the lost depth also is the reason why 3D location labels are the most important and difficult part for monocular 3D detection.
	LiDAR point clouds, which provide the crucial 3D measurements, are indispensable in the labeling procedure.
	As a common practice, annotators annotate 3D boxes on the LiDAR points clouds.
	On the other hand, concerning the data collecting process in a self-driving system, a large number of successive snippets are collected.	
	Generally speaking, to save the high annotation costs, only some key frames in collected snippets are labeled to train networks, such as KITTI dataset \cite{KITTI2012}.
	Consequently, massive LiDAR point clouds holding valuable 3D information remain unlabeled.

	Inspired by the 3D location label requirement and accurate LiDAR 3D measurements in 3D space, we propose a general and intuitive framework to make full use of LiDAR point clouds, dubbed {\bf LPCG} (LiDAR point cloud guided monocular 3D object detection).
	Specifically, we use unlabeled LiDAR point clouds to generate pseudo labels, converting unlabeled data to training data for monocular 3D detectors.
	These pseudo labels are not as accurate as the manually annotated labels, but they are good enough for the training of monocular 3D detectors due to accurate LiDAR 3D measurements.

	We further present two working modes in LPCG: the high accuracy mode and the low cost mode, to generate different qualities of pseudo labels according to annotation costs. 
	The high accuracy mode requires a small amount of labeled data to train a LiDAR-based detector, and then the trained detector can produce high-quality pseudo labels on other unlabeled data.
	This manner can largely boost the accuracy of monocular 3D detectors.
	Additionally, we propose a heuristic method to produce pseudo labels without requiring any 3D box annotation.
	Such pseudo labels are directly obtained from the RoI LiDAR point clouds, by employing point clustering and minimum bounding box estimation. 
	We call this manner the low cost mode.
	Either the high accuracy mode or the low cost mode in LPCG can be plugged into any monocular 3D detector.
	
	Based on the above two modes, we can fully use LiDAR point clouds, allowing monocular 3D detectors to learn desired objectives on a large training set meanwhile avoiding architecture modification and removing extra annotation costs.
	By applying the framework (high accuracy mode), we significantly increase the 3D and BEV (bird's-eye-view)  AP of prior state-of-the-art methods \cite{MonoGRNet,M3D,RTM3D,GrooMeD-NMS,MonoFlex}.
	In summary, our contributions are two folds as below:

\begin{itemize}
		\item {\bf First}, we analyze requirements in terms of the label accuracy towards the training of monocular 3D detection.
		 Based on this analysis, we introduce a general framework that can utilize massive unlabeled LiDAR point clouds, to generate new training data with valuable 3D information for monocular methods during the training.
		\item {\bf Second}, experiments show that the baseline detector employing our method outperforms recent SOTA methods by a large margin, ranking $1^{st}$ on KITTI \cite{KITTI2012} monocular 3D and BEV detection benchmark at the time of submission (car, March. 2022). 
		In Waymo \cite{waymo} benchmark, our method achieves close accuracy compared to the baseline detector using 100\% labeled data while our method requires only 10\% labeled data with 90\% unlabeled data. 
\end{itemize}

\section{Related Work}
\subsection{LiDAR-based 3D Object Detection}
	The LiDAR device can provide accurate depth measurement of the scene,  thus is employed by most state-of-the-art 3D object detection methods \cite{PART,PP,Point-gnn,3dssd,HVNet,PVRCNN}. 
	These methods can be roughly divided into voxel-based methods and point-based methods. 
	Voxel-based methods \cite{VoxelNet} first divide the point cloud into a voxel grid and then feed grouped points into fully connected layers, constructing unified feature representations.
	They then employ 2D CNNs to extract high-level voxel features to predict 3D boxes.
	By contrast, point-based methods \cite{Pointrcnn,FPointnet} directly extract features on the raw point cloud via fully connected networks, such as PointNet \cite{Pointnet} and PointNet++ \cite{Pointnet++}.
	SOTA 3D detection methods predominantly employ LiDAR point clouds both in training and inference, while we only use LiDAR point clouds in the training stage.

\subsection{Image-only-based Monocular 3D Object Detection}
	As a commonly available and cheap sensor, the camera endows 3D object detection with the potential of being adopted everywhere.
	Thus monocular 3D object detection has become a very popular area of research and has developed quickly in recent years. 
	Monocular works can be categorized into image-only-based methods \cite{M3D,RTM3D} and depth-map-based methods \cite{PseudoLidar,AM3D,PatchNet} according to input representations.
	M3D-RPN \cite{M3D} employs different convolution kernels in row-spaces that can explore different features in specific depth ranges and improve 3D estimates with the 2D-3D box consistency.
	Furthermore,  RTM3D \cite{RTM3D} predicts perspective key points and initial guesses of objects' dimensions/orientations/locations, where the key points are further utilized to refine the initial guesses by solving a constrained optimization problem. 
	More recently, many image-only-based works utilize depth estimation embedding \cite{MonoFlex}, differentiable NMS \cite{GrooMeD-NMS}, and geometry properties \cite{MonoEF,Ground-Aware,GUPNet}, obtaining great success. 
	There is also a related work \cite{SDF} that introduces a novel autolabeling strategy of suggesting a differentiable template matching model with curriculum learning, using differentiable rendering of SDFs, while the pipeline is rather complicated.

			\begin{figure*}[t]
		\begin{center}
				\includegraphics[width=1.0\linewidth]{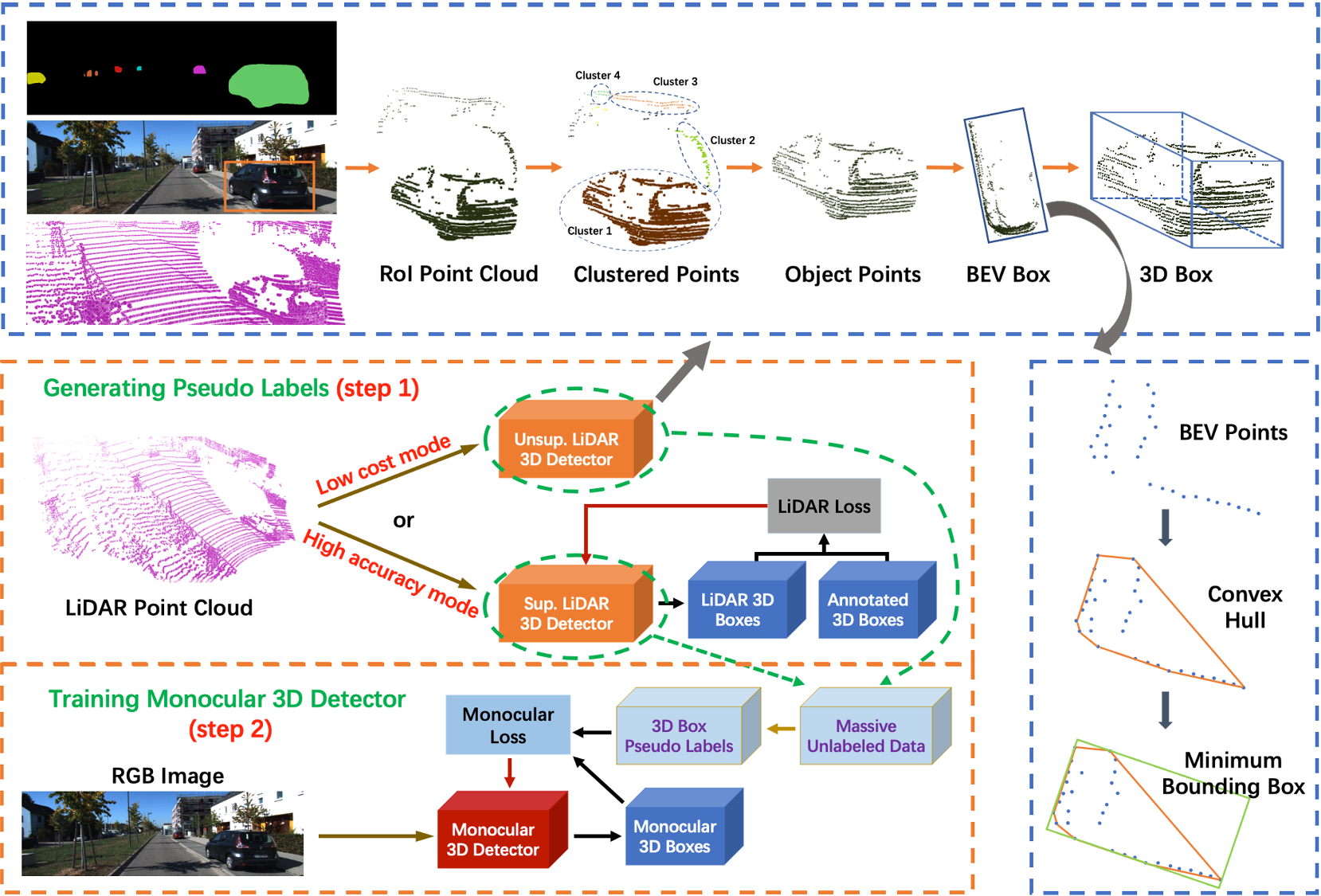}
		\end{center}
		\caption{
		Overview framework. We generate 3D box pseudo labels from unlabeled LiDAR point clouds, aiming to train the monocular 3D detector. Such 3D boxes are predicted via the well-trained LiDAR 3D detector (high accuracy mode) or obtained directly from the point cloud without training (low cost mode).
		``Unsup." and ``Sup." in the figure denote unsupervised and supervised, respectively.
		}
		\label{fig: over_framework}
	\end{figure*}

\subsection{Depth-map-based Monocular 3D Object Detection}
	Although monocular methods are developing quickly, a large performance gap still exists compared to LiDAR-based methods.
	Some prior works \cite{PseudoLidar,Mono_PseudoLidar} argue that the improper choice of data representation is one of the main reasons, and propose to use transformed image-based depth maps.
	They first project LiDAR point clouds onto the image plane, to form depth map labels to train a depth estimator.
	Pseudo-LiDAR \cite{PseudoLidar} converts the image into a point cloud by using an estimated depth map and then conducts 3D detection on it.
	They show promising results compared to previous image-only-based methods.
	Inspired by this, many later methods \cite{Mono_PseudoLidar,AM3D,D4LCN,PatchNet} also utilize off-the-shelf depth estimates to aim 3D detection and gain performance improvements.
	More recently, CaDDN \cite{CaDDN} integrates the dense depth estimation into monocular 3D detection, by using a predicted categorical depth distribution to project contextual features to the 3D space.
	Compared to previous depth-map-based methods, we aim to explore the potential of using LiDAR point clouds to generate pseudo labels for monocular 3D detectors.

\section{LiDAR Guided Monocular 3D Detection}
	In this section, we detail the proposed framework, namely, LPCG (LiDAR Guided Monocular 3D Detection).
	First, as shown in Figure \ref{fig: dis}, the manually annotated perfect labels are unnecessary for monocular 3D detection.
	The accuracy led by disturbed labels (5\%) is comparable to the one led by perfect labels.
	When enforcing large disturbances (10\% and 20\%), we can see that the location dominates the performance (the AP dramatically decreases only when disturbing the location). 
	It indicates that rough pseudo 3D box labels with precise locations may replace the perfect annotated 3D box labels.

	We note that LiDAR point clouds can provide valuable 3D location information.
	More specifically, LiDAR point clouds provide accurate depth measurement within the scene, which is crucial for 3D object detection as precise surrounding depths indicate locations of objects. 
	Also, LiDAR point clouds can be easily captured by the LiDAR device, allowing a large amount of LiDAR point clouds to be collected offline without manual cost.
	Based on the analysis above, we use LiDAR point clouds to generate 3D box pseudo labels.
	The newly-generated labels can be used to train monocular 3D detectors.
	This simple and effective way allows monocular 3D detectors to learn desired objectives meanwhile eliminating annotation costs on unlabeled data.
	We show the overall framework in Figure \ref{fig: over_framework}, in which the method is able to work in two modes according to the reliance on 3D box annotations.
	If we use a small amount of 3D box annotations as prior, we call it the high accuracy mode since this manner leads to high performances.
	By contrast, we call it the low cost mode if we do not use any 3D box annotation.

\subsection{High Accuracy Mode}{\label{sec:sup}}
	To take advantage of available 3D box annotations, as shown in Figure \ref{fig: over_framework}, we first train a LiDAR-based 3D detector from scratch with LiDAR point clouds and associated 3D box annotations.
	The pre-trained LiDAR-based 3D detector is then utilized to infer 3D boxes on other unlabeled LiDAR point clouds.
	Such results are treated as pseudo labels to train monocular 3D detectors. 
	We compare the pseudo labels with manually annotated perfect labels in Section \ref{sec:comare}. 
	Due to precise 3D location measurements, pseudo labels predicted from the LiDAR-based 3D detector are rather accurate and qualified to be used directly in the training of monocular 3D detectors.
	We summarize the outline in Algorithm \ref{algo_acc}.

\begin{algorithm}[h]
\textbf{Input:} Labeled data $A:\{A_{data},A_{label}\}$, unlabeled data $B:\{B_{data}\}$\\
\textbf{Output:} Well-trained monocular 3D detection model $M_{mono}$ \\
\SetAlgoLined
$M_{lidar}$ $\leftarrow$ Training a supervised LiDAR-based 3D detection model on labeled data $\{A_{data},A_{label}\}$. \\

$\{B_{pseudo-label}\}$ $\leftarrow$ Conducting predictions from LiDAR point clouds on unlabeled data: $M_{lidar}(B_{data})$ \\

 $C:\{C_{data},C_{label}\}$ $\leftarrow$ Merging training data:  $\{A_{data}\cup B_{data},A_{label}\cup B_{pseudo-label}\}$ \\

 $M_{mono}$ $\leftarrow$ Training a supervised monocular-based model on new set $\{C_{data},C_{label}\}$. \\

Return  $M_{mono}$ \\
 \caption{Outline of  the high accuracy mode in LPCG. Both labeled and unlabeled training data contains RGB images and associated LiDAR point clouds.}
 \label{algo_acc}
\end{algorithm}

	Interestingly, with different training settings for the LiDAR-based 3D detector, we empirically find that monocular 3D detectors trained by resulting pseudo labels show close performances.  
	It indicates that monocular methods can indeed be beneficial from the guidance of the LiDAR point clouds and only a small number of 3D box annotations are sufficient to push the monocular method to achieve high performance. 
	Thus the manual annotation cost of high accuracy mode is much lower than the one of the previous manner.
	Detailed experiments can be found in Section \ref{sec:abla}.
	Please note, the observations on label requirements and 3D locations are the core motivation of LPCG.
	The premise that LPCG can work well is that LiDAR points provide rich and precise 3D measurements, which offer accurate 3D locations.

\subsection{Low Cost Mode}{\label{sec:unsup}}
	In this section, we describe  the method of using LiDAR point clouds to eliminate the reliance on manual 3D box labels. 	
	First, an off-the-shelf 2D instance segmentation model \cite{mask-rcnn} is adopted to perform segmentation on the RGB image, obtaining 2D box and mask estimates.
	These estimates are then used for building camera frustums in order to select associated LiDAR RoI points for every object, where those boxes without any LiDAR point inside are ignored.
	However, LiDAR points located in the same frustum consist of object points and mixed background or occluded points.
	To eliminate irrelevant points, we take advantage of DBSCAN \cite{DBSCAN} to divide the RoI point cloud into different groups according to the density. 
	Points that are close in 3D spatial space will be aggregated into a cluster. 
	We then regard the cluster containing most points as a target corresponding to the object. 
	Finally, we seek the minimum 3D bounding box that covers all target points.

	To simplify the problem of solving the 3D bounding box, we project points onto the bird's-eye-view map, reducing parameters since the height ($h$) and $y$ coordinate (under camera coordinate system) of the object can be easily obtained. 
	Therefore, we have:
\begin{equation}
L=\mathop{\rm min}\limits_{B_{bev}}(Area(B_{bev})), \quad \textit{subject to $p$ is inside $B_{bev}$,  where $p$ $\in LiDAR_{RoI}$}
\label{dis_error}
\end{equation}
	where $B_{bev}$ refers to a bird's-eye-view (BEV) box. 
	We solve this problem by using the convex hull of object points followed by obtaining the box by using rotating calipers \cite{Calipers}.
	Furthermore, the height $h$ can be represented by the max spatial offset along the y-axis of points, and the center coordinate $y$ is calculated by averaging $y$ coordinates of points.
	We use a simple rule of restricting object dimensions to remove outliers.
	The overall training pipeline for monocular methods is summarized in Algorithm \ref{algo_cost}.

	\begin{algorithm}[h]
\textbf{Input:} Unlabeled data $D:\{D_{data-image},D_{data-lidar}\}$, pre-trained Mask-RCNN model: $M_{mask}$\\
\textbf{Output:} Well-trained monocular 3D detection model $M_{mono}$ \\
\SetAlgoLined
$Mask_{2D}$ $\leftarrow$ Conducting predictions from RGB images: $M_{mask}(D_{data-image})$. \\

$LiDAR_{RoI}$ $\leftarrow$ Selecting  and clustering LiDAR point clouds from $D_{data-lidar}$ by $Mask_{2D}$\\

 $D_{pseudo-label}$ $\leftarrow$ Gernerating pseudo labels on RoI LiDAR points $LiDAR_{RoI}$. \\
 
 $M_{mono}$ $\leftarrow$ Training a supervised monocular-based model on new data $\{D_{data-image},D_{pseudo-label}\}$. \\

Return  $M_{mono}$ \\
 \caption{
 Outline of  the low cost mode in LPCG. Unlabeled data contains RGB images and associated LiDAR point clouds.
 }
 \label{algo_cost}
\end{algorithm}

\begin{table}[h]
      \centering
      \scriptsize
      			\caption{
			Comparisons of different modes in previous works and ours. 
			}
			\label{tab:com}
			\begin{tabular}{l|c|c|c}
				\toprule   
				Approaches &  Modality & 3D box annotations & Unlabeled LiDAR data \\ 
				\midrule         
				\multirow{2}{*}{Previous} & Image-only based  & Yes & No \\ 
				~& Depth-map based  & Yes & Yes \\ 
				\midrule
				\multirow{2}{*}{Ours} & High accuracy mode & Yes & Yes \\ 
				~& Low cost mode  & No & Yes \\ 
				\bottomrule
			\end{tabular}
  \end{table}

	\begin{figure}[h]
		\begin{center}
				\includegraphics[width=0.85\linewidth]{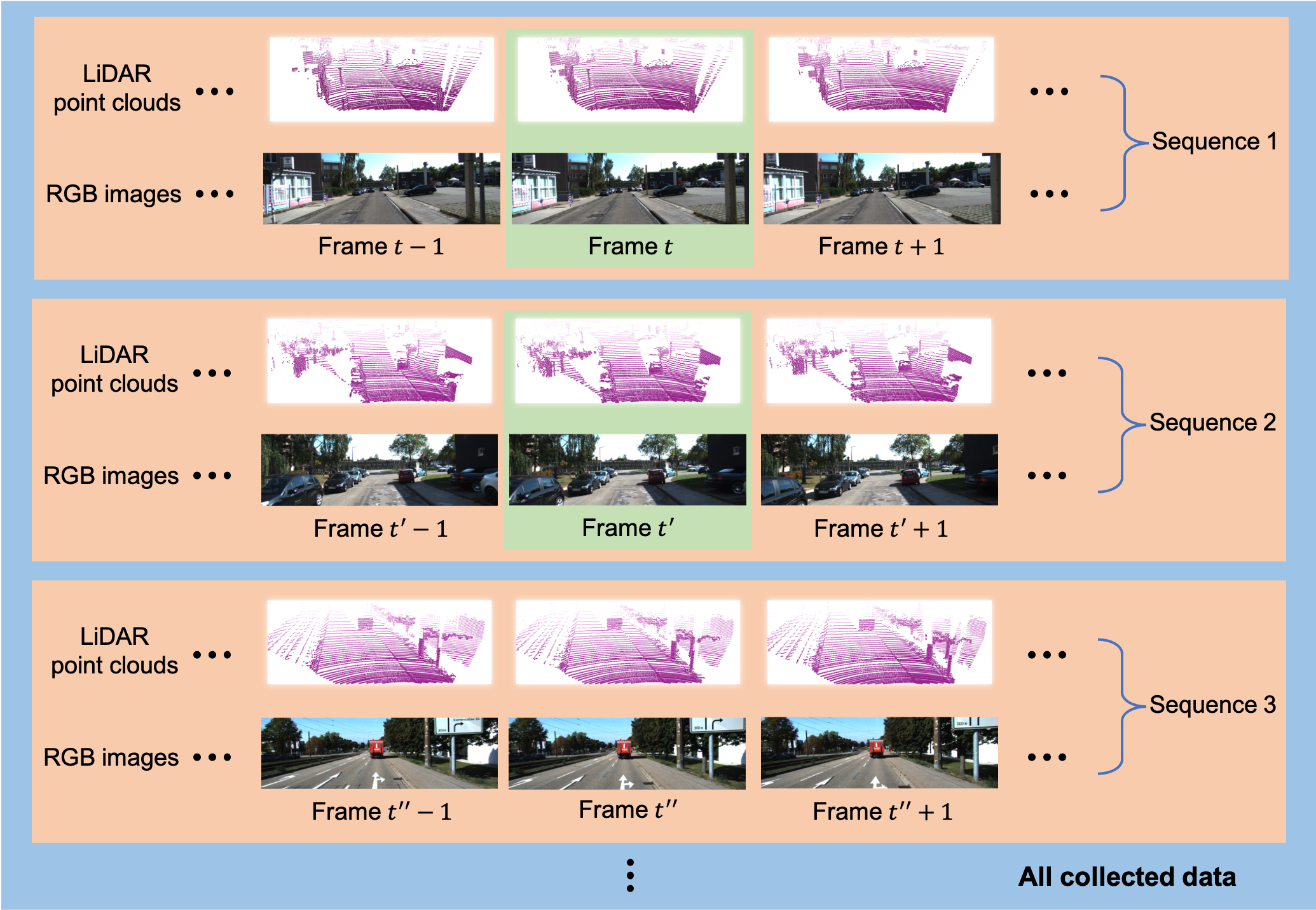}
		\end{center}
		\caption{
		Data collecting strategy in a real-world system.
		Only some sequences ({\it{e.g.}}, sequence 1 and 2) are chosen for annotating because of the limited time and resources in the real world, such as Waymo \cite{waymo}.
		Further, concerning the high annotation costs, only some key frames ({\it{e.g.}}, frame $t$ and $t^\prime$) in the selected sequences are annotated, such as KITTI \cite{KITTI2012}.
		} 
		\label{fig: data_colloct}
	\end{figure}

\section{Applications in Real-world Self-driving System}
	In this section, we describe the application of LPCG to a real-world self-driving system.
	First, we illustrate the data collecting strategy in Figure \ref{fig: data_colloct}.
	Most self-driving systems can easily collect massive unlabeled LiDAR point cloud data and synchronized RGB images.
	This data is organized with many sequences, where each sequence often refers to a specific scene and contains many successive frames.
	Due to the limited time and resources in the real world, only some sequences are chosen for annotating, to train the network, such as Waymo \cite{waymo}.
	Further, to reduce the high annotation costs, only some key frames in the selected sequences are annotated, such as KITTI \cite{KITTI2012}.
	Therefore, there remains massive unlabeled data in real-world applications.

	Considering that LPCG can fully take advantage of the unlabeled data, it is natural to be employed in a real-world self-driving system.
	Specifically, the high accuracy mode only requires a small amount of labeled data.
	Then we can generate high-quality training data from remaining unlabeled data for monocular 3D detectors, to boost the accuracy.
	In experiments, we quantitatively and qualitatively show that the generated 3D box pseudo labels are good enough for monocular 3D detectors.
	Additionally, the low cost mode does not require any 3D box annotation, still providing accurate 3D box pseudo labels.
	We compare LPCG with previous methods in Table \ref {tab:com} in terms of the data requirements.

		  \begin{table}[h]
	  \scriptsize
	  \setlength{\tabcolsep}{1.5mm}
      \centering
      			\caption{Comparisons on KITTI testing set. 
			We use \textcolor{red}{red} to indicate the highest result and \textcolor{blue}{blue} for the second-highest result and \textcolor{cyan}{cyan} for the third-highest result.
			$\textcolor[rgb]{0.00,0.00,1.00}{^\ddagger}$ denotes the baseline detector we employed, and the improvements are relative to the baseline detectors.
			We define the new state of the art.
			Please note, DD3D\cite{DD3D}$\textcolor[rgb]{0.00,0.00,1.00}{^\star}$ employs both the large private DDAD15M dataset (containing approximately 15M frames) and the KITTI depth dataset (containing approximately 26K frames).
			}
			\label{tab:test}
			\begin{tabular}{l|c|p{1.8cm}<{\centering}p{1.8cm}<{\centering}p{1.8cm}<{\centering}}
				\toprule   
				\multirow{2}{*}{Approaches} &  \multirow{2}{*}{Extra data} & \multicolumn{3}{c}{AP$_{BEV}$/AP$_{3D}|$ (IoU=0.7)$|\scriptstyle R_{40}$} \\ 
				~ & ~& Easy & Moderate & Hard\\ 
				\midrule         
				ROI-10D \cite{ROI10D} &KITTI depth & 9.78/4.32  & 4.91/2.02  & 3.74/1.46 \\ 
				MonoGRNet \cite{MonoGRNet} &None& 18.19/5.74  & 11.17/9.61  & 8.73/4.25 \\
				AM3D \cite{AM3D}&KITTI depth& 25.03/16.50 & {17.32}/10.74 & {{14.91/9.52}} \\
				MonoPair \cite{MonoPair} &None& 19.28/13.04 & 14.83/9.99 & 12.89/8.65 \\
				D4LCN \cite{D4LCN}  &KITTI depth& 22.51/{{16.65}} & 16.02/{11.72} & 12.55/9.51 \\
				RTM3D \cite{RTM3D} &None& 19.17/14.41  & 14.20/10.34  &  11.99/8.77    \\ 
				PatchNet \cite{PatchNet}&KITTI depth& 22.97/15.68 & 16.86/{{11.12}} & {14.97/10.17} \\
				Neighbor-Vote \cite{Neighbor-Vote} &KITTI depth & 27.39/15.57 & 18.65/9.90 & 16.54/8.89  \\
				MonoRUn \cite{MonoRUn}  &None& 27.94/19.65 & 17.34/12.30 & 15.24/10.58  \\
				MonoRCNN \cite{MonoRCNN}  &None&  25.48/18.36 & 18.11/12.65 & 14.10/10.03  \\
				Monodle \cite{monodle} &None & 24.79/17.23 & 18.89/12.26 & 16.00/10.29  \\
				CaDDN \cite{CaDDN} &None & 27.94/19.17 & 18.91/13.41 & 17.19/11.46  \\
				Ground-Aware \cite{Ground-Aware}&None  & 29.81/21.65& 17.98/13.25 & 13.08/9.91  \\
				GrooMeD-NMS \cite{GrooMeD-NMS} &None& 26.19/18.10 & 18.27/12.32 & 14.05/9.65\\
				MonoEF \cite{MonoEF}&None  & 29.03/21.29 & 19.70/13.87 & {\bf\textcolor{cyan}{17.26}}/11.71  \\
				DDMP-3D \cite{DDMP-3D}  &KITTI depth& 28.08/19.71 & 17.89/12.78 & 13.44/9.80  \\
				PCT \cite{PCT} &KITTI depth & 29.65/21.00 & 19.03/13.37 & 15.92/11.31 \\
				AutoShape \cite{Autoshape} &None & \bf \textcolor{cyan}{30.66/22.47} & 20.08/14.17 & 15.95/11.36  \\
				GUPNet \cite{GUPNet}  &None& 30.29/22.26 & \bf \textcolor{blue}{21.19/15.02} & \bf \textcolor{blue}{18.20/13.12}  \\
				M3D-RPN \cite{M3D} $\textcolor[rgb]{0.00,0.00,1.00}{^\ddagger}$ &None & 21.02/14.76  & 13.67/9.71  & 10.23/7.42   \\
				MonoFlex \cite{MonoFlex} $\textcolor[rgb]{0.00,0.00,1.00}{^\ddagger}$ &None  & 28.23/19.94 & 19.75/13.89 & 16.89/12.07  \\
				\textit{DD3D \cite{DD3D} $\textcolor[rgb]{0.00,0.00,1.00}{^\star}$}  &\textit{DDAD15M...} &{\textit{32.35/23.19}} &{\textit{23.41/16.87}} & {\textit{20.42/14.36}}  \\
				\midrule 
				{\bf LPCG+M3D-RPN\cite{M3D}} &\multirow{2}{*}{KITTI depth} & {\bf \textcolor{blue}{30.72/22.73}} & {\bf \textcolor{cyan}{20.17/14.82}} &   16.76/{\bf \textcolor{cyan}{12.88}}\\
				 Improvements (to baseline) & ~ & +9.70/+7.97  & +6.50/+5.11  &  +6.53/+5.46\\ 
				\midrule
				{\bf LPCG+MonoFlex\cite{MonoFlex}}  &\multirow{2}{*}{KITTI depth} & \bf \textcolor{red}{35.96/25.56}  & \bf \textcolor{red}{24.81/17.80}  &\bf \textcolor{red}{21.86/15.38}  \\
				Improvements (to baseline) &~  & +7.73/+5.62  & +5.06/+3.91  & +4.97/+3.31  \\ 
				\bottomrule 
			\end{tabular}
  \end{table}

	\section{Experiments}
\subsection{Implementation Details}
	We use the image-only-based monocular 3D detector M3D-RPN \cite{M3D}, and adopt PV-RCNN \cite{PVRCNN} as the LiDAR 3D detector for the high accuracy mode.
	We filter out 3D boxes generated from LiDAR point clouds with the confidence of 0.7.
	Experiments on other methods are conducted by the official code that is publicly available, and all settings keep the same as the original paper.
	 During the process of using LiDAR point clouds to train monocular 3D detectors,  the learning iteration is scaled according to the number of training data.
	The high accuracy mode is employed by default.
	For the low cost mode, we use Mask-RCNN \cite{mask-rcnn} pre-trained in the COCO dataset \cite{COCO}, and filter the final 3D bounding box by the width range of 1.2-1.8 meters and the length range of 3.2-4.2 meters.
	We filter out 2D boxes predicted from Mask-RCNN \cite{mask-rcnn} with the confidence of 0.9.
	More details and ablations are provided in the supplementary material as the space limitation.

\subsection{Dataset and Metrics}
\noindent
	{\bf Dataset.} 
	Following prior works \cite{MonoGRNet,AP40,M3D,PseudoLidar,RTM3D,PatchNet}, experiments are conducted on the popular KITTI 3D object dataset \cite{KITTI2012}, which contains 7, 481 manually annotated images for training and 7, 518 images for testing.  
	Due to groundtruths of the test set are not available, the public training set is further split into two subsets  \cite{3DOP}: training set (3, 712 images) and validation set (3, 769 images). 
	Following the fashion, we report our results both on the validation set and the test set.
	And we use the validation set for all ablations.
	Also, our method and depth-map-based methods use RGB images and synchronized LiDAR point clouds from KITTI raw scenes.
	For depth-map-based methods,  note that the original depth training set overlaps KITTI 3D detection validation set. 
	Therefore we exclude scenes that emerge in KITTI 3D validation set to avoid data leakage \cite{Are_we,OCM3D}.
	LiDAR point clouds in the remaining scenes are used.
	We call this extra dataset the KITTI depth dataset.
	It provides approximately 26K samples to train the depth estimator (for most depth-map-based methods) or to generate extra training samples for monocular 3D detectors (LPCG).
	
	Additionally, to further validate the effectiveness of LPCG, we conduct experiments on the Waymo Open Dataset \cite{waymo}, which is a modern large dataset.
	 It contains 798 training sequences and 202 validation sequences, and we adopt the same data processing strategy proposed in CaDDN \cite{CaDDN}.
	 The sampled training dataset includes approximately 50K training samples with manual annotations.

 ~\\
 \noindent
	{\bf Metrics.} 
	Each manually annotated object is divided into easy, moderate, and hard levels according to the occlusion, truncation, and 2D box height \cite{KITTI2012}.
	Average precisions (AP) on the car class for bird's-eye-view (BEV) and 3D boxes with 0.5/0.7 IoU thresholds are commonly used metrics for monocular 3D detection.
	Many previous methods utilize the $AP_{11}$ metric, which has an overrated issue \cite{AP40},  and $AP_{40}$ \cite{AP40} is proposed to resolve it. 
	We report $AP_{40}$ results to make comprehensive comparisons.
	For Waymo dataset, we adopt the official mAP and mAPH metrics.

	  \begin{table}[h]
	  \setlength{\tabcolsep}{1.5mm}
      \centering
      \scriptsize
      			\caption{
			Comparisons with SDFLabel \cite{SDF}. Note that here we use the same number of training samples for fair comparisons.
			}
			\label{tab:unsup}
			\begin{tabular}{l|c|ccc}
				\toprule   
				\multirow{2}{*}{Approaches} & \multirow{2}{*}{Data requirements in training}& \multicolumn{3}{c}{AP$_{3D}$ (IoU=0.7)$|\scriptstyle R_{40}$} \\
				~  & ~&Easy & Moderate & Hard \\ 
				\midrule         
				SDFLabel \cite{SDF} & 2D masks+LiDAR+CAD models & 1.23 & 0.54 & -\\
				\midrule 
				LPCG (low cost mode) & 2D masks+LiDAR& \bf  5.36 & \bf 3.07 & \bf 2.32\\
				\bottomrule 
			\end{tabular}
  \end{table}

\subsection{Results on KITTI}
	We evaluate LPCG on KITT test set using two base monocular detectors\cite{M3D,MonoFlex} with the high accuracy mode. 
	Table \ref{tab:test} shows quantitative results in {\it{test}} set.
	Due to the space limitation, qualitative results are included in the supplementary material.
	We can observe that our method increases the current SOTA BEV/3D AP from  {\bf 21.19/15.02} to {\bf 24.81/17.80} under the moderate setting, which is rather significant.
	Even using a monocular detector \cite{M3D} proposed in 2019, our method still allows it to achieve new state-of-the-art compared to prior works.
	Note that our method still performs better, while DD3D \cite{DD3D} employs both the large private DDAD15M dataset (containing approximately 15M frames) and the KITTI depth dataset (containing approximately 20K frames). 
	Also, we boost the performance on pedestrian and cyclist categories of the original method, and provide the results in Table \ref{tab:ped_cyc}.
	Such results prove the effectiveness of LPCG.

	For the low cost mode, we note that there are few works exploring this area, namely, few works have explored monocular 3D detection without any 3D box annotation.
	The most related work is SDFLabel \cite{SDF}, which also does not require 3D box annotation.
	Thus we compare LPCG with the low cost mode with SDFLabel \cite{SDF} in Table \ref{tab:unsup}.
	Please note, in this experiment we use the same number of training samples, namely, the 3, 769 samples in KITTI3D training set.
	Our method outperforms it by a large margin, and our method is more generally usable as our pipeline is much simpler than SDFLabel \cite{SDF}.

	 		\begin{table*}[h]
        \centering
        \scriptsize
        			        \caption{
			        Comparisons on Waymo. ``lab." and ``unlab." denote labeled and unlabeled.
			        	 }
			\label{tab:waymo}   
			\begin{tabular}{c|c|c|cccc}
				\toprule  
				Difficulty &w/ LPCG & Data requirements & Overall & 0$-$30m & 30$-$50m & 50m$-$$\infty$ \\   
				\midrule
				\multicolumn{7}{c}{{\textit{\textbf{under 3D mAP metric}}}}    \\
				\midrule
				\multirow{3}{*}{\tabincell{c}{LEVEL 1\\(IOU=0.5)}}& No & 10\% labeled data  &  4.14 &14.64  & 1.63  & 0.04    \\ 
				~ & No & 100\% labeled data  &  6.42 & 19.50  & 3.04  & 0.17    \\ 
				~&  Yes & 10\% lab. + 90\% unlab. data   & 6.23   & 18.39  &  3.44   & 0.19  \\ 
				 				\midrule     								
				\multirow{2}{*}{\tabincell{c}{LEVEL 2\\(IOU=0.5)}}& No & 10\% labeled data &  3.88   & 14.59  &1.58   & 0.04 \\ 
				~ & No & 100\% labeled data &  6.02   & 19.43  &2.95   & 0.15 \\ 
				~& Yes & 10\% lab. + 90\% unlab. data  & 5.84   & 18.33  &  3.34   & 0.17  \\ 
				\midrule
				 \multicolumn{7}{c}{{\textit{\textbf{under 3D mAPH metric}}}}    \\
				\midrule
				\multirow{3}{*}{\tabincell{c}{LEVEL 1\\(IOU=0.5)}}& No & 10\% labeled data  & 3.94 & 14.07 & 1.51 & 0.04\\ 
				~ & No & 100\% labeled data  & 6.19 & 18.88 & 2.89 & 0.16\\ 
				~& Yes & 10\% lab. + 90\% unlab. data  & 6.09& 18.03 &  3.33 &  0.17\\ 
				 				\midrule     								
				\multirow{3}{*}{\tabincell{c}{LEVEL 2\\(IOU=0.5)}}& No & 10\% labeled data  & 3.69& 14.02 & 1.46 & 0.03\\ 
				~& No & 100\% labeled data  & 5.80 & 18.81 &2.80 & 0.14\\ 
				~& Yes & 10\% lab. + 90\% unlab. data  &  5.70 & 17.97 &  3.23 &  0.15\\ 
				\bottomrule 
			\end{tabular}
      \end{table*}

		  \begin{table}[h]
      \centering
      \scriptsize
      			\caption{Improvements on other categories.
			}
			\label{tab:ped_cyc}
			\begin{tabular}{l|p{1.2cm}<{\centering}p{1.2cm}<{\centering}p{1.2cm}<{\centering}|p{1.2cm}<{\centering}p{1.2cm}<{\centering}p{1.2cm}<{\centering}}
				\toprule   
				 \multirow{2}{*}{Approaches}  & \multicolumn{3}{c|}{Pedestrian, AP$_{3D}$ (IoU=0.5)$|\scriptstyle R_{40}$} & \multicolumn{3}{c}{Cyclist, AP$_{3D}$ (IoU=0.5)$|\scriptstyle R_{40}$} \\
				~ & Easy & Moderate & Hard & Easy & Moderate & Hard \\ 
				\midrule         
				M3D-RPN \cite{M3D} & 4.75 & 3.55 & 2.79 & 3.10 & 1.49 & 1.17\\
				\bf M3D-RPN+LPCG  & \bf 7.21 & \bf 5.53 & \bf 4.46& \bf 4.83 & \bf 2.65 & \bf 2.62\\
				\bottomrule 
			\end{tabular}
  \end{table}

\subsection{Results on Waymo}{\label{sec:gap}}
	To further prove the effectiveness of our method, we conduct experiments on the Waymo open dataset.
	Concerning its large scale, when enough perfect labels are available, in this dataset we aim to investigate the performance gap between the generated pseudo labels and the manual 3D box annotations.
	More specifically, we use the baseline detector M3D-RPN \cite{M3D}, training it with pseudo labels and manual annotations, respectively.
	We report the results in Table \ref{tab:waymo}. 
	Pseudo labels on unlabeled data are generated by the high accuracy mode in LPCG. 
	Interestingly, we can see that the detector using 10\% labeled data and 90\% unlabeled data achieves comparable accuracy to the one using 100\% labeled data ({\it{e.g.}}, 6.42 {\it{vs.}} 6.23 and 6.19 {\it{vs.}} 6.09).
	This result demonstrates the generalization ability of LPCG, indicating that LPCG can also reduce the annotation costs for the large scale dataset with slight accuracy degradation.

\subsection{Comparisons on Pseudo Labels and Manually Annotated Labels }{\label{sec:comare}}
	As expected, pseudo labels are not as accurate as manually annotated labels. 
	It is interesting to quantitatively evaluate pseudo labels using manually annotated labels.
	We report the results in Table \ref{tab:com_pseudo_gt}.
	TP, FP, FN are calculated by matching pseudo labels and annotated labels.
	Regarding matched objects, we average the relative error (MRE) on each group of 3D box parameters (location, dimension, and orientation).
	We can see that pseudo labels from the high accuracy mode can match most real objects (91.39\%), and the mean relative errors are 1\%-6\%.
	Therefore pseudo labels from the high accuracy mode are good enough for monocular 3D detectors.
	Actually, experiments in Table \ref{tab:test} and \ref{tab:waymo} also verify the effectiveness.
	On the other hand, for the low cost mode, we can see that many real objects are missed (11834).
	We note that missed objects are often truncated, occluded, or faraway.
	The attached LiDAR points cannot indicate the full 3D outline of objects, thus they are hard to recover by geometry-based methods.

\begin{table}[h]
      \centering
      \scriptsize
      			\caption{
			Performance of pseudo labels on \textit{val} set.
			We evaluate pseudo labels using manually annotated labels.
			``MRE" refers to the mean relative error ({\it{e.g.}}, the relative error of location is $\frac{Error_{Loc}}{Loc}$). 
		``Loc., Dim., Orient." are the location ($x,y,z$), dimension ($h,w,l$), and orientation ($R_y$).
			``TP, FP, FN" are the true positive, false positive, and false negative, which are calculated by matching pseudo labels and annotated labels.
			Please see Section \ref{sec:comare} for detailed analysis. 
			Note that pseudo labels on \textit{val} set are just for the evaluation, and they are not used in the training of monocular 3D detectors.
			}
			\label{tab:com_pseudo_gt}
			\begin{tabular}{l|ccc|p{2.0cm}<{\centering}|p{2.0cm}<{\centering}|c}
				\toprule   
				Pseudo label types & TP & FP &  FN & Loc. MRE & Dim. MRE & Orient. MRE \\
				\midrule         
				Low cost mode & 2551 & 161 & 11834 &4\%/5\%/2\% & 8\%/6\%/7\% &8\%  \\
				High accuracy mode& 13146 & 3299 & 1239 & 4\%/4\%/1\% & 4\%/4\%/6\% & 4\%\\
				\bottomrule 
			\end{tabular}
\end{table}

			\begin{table*}[h]
			\begin{center}
			\scriptsize
					\caption{ 
		Ablation for annotation numbers. 
		3712 is the total annotations in the KITTI 3D training dataset.
		All results are evaluated on KITTI {\it{val}} set with metric $AP|_{R_{40}}$.
		}
		\label{tab:num_anno}
			\begin{tabular}{l|p{2.0cm}<{\centering}p{2.0cm}<{\centering}p{2.0cm}<{\centering}}
				\toprule  
				\multirow{2}{*}{Annotations} &  \multicolumn{3}{c}{AP$_{BEV}$/AP$_{3D}|$ (IoU=0.7)$\scriptstyle R_{40}$}  \\ 
				 ~ & Easy & Moderate & Hard\\ 
				\midrule
				100  &  30.19/20.90  & 21.96/15.37  & 19.16/13.00 \\ 
				200   &  30.39/22.55  & 22.44/16.17  & 19.60/14.32\\ 
				500  &  32.01/23.13  & 23.31/17.42  & 20.26/14.95\\ 
				1000 &  33.08/25.71  & 24.89/19.29 & 21.94/16.75\\ 
				\midrule
				 3712  & \bf 33.94/26.17 &\bf 25.20/19.61   &\bf 22.06/16.80\\
				\bottomrule 
			\end{tabular}
		\end{center}
	\end{table*}

\begin{table*}[h]
		\begin{center}
		\scriptsize
				\caption{Extension on different monocular detectors. LPCG can be easily plugged into other methods. $^*$ denotes that the model is re-implemented by us. All the methods are evaluated on KITTI {\it{val}} set with metric $AP|_{R_{40}}$.}
		\label{tab:extension_3d}
			\begin{tabular}{l|ccc|ccc}
				\toprule  
				\multirow{2}{*}{Approaches} &  \multicolumn{3}{c|}{AP$_{3D}$ (IoU=0.5)$|\scriptstyle R_{40}$}  &  \multicolumn{3}{c}{AP$_{3D}$ (IoU=0.7)$|\scriptstyle R_{40}$} \\ 
				~ &  Easy & Moderate & Hard & Easy & Moderate & Hard\\ 
				\midrule             
				MonoGRNet \cite{MonoGRNet} &  47.34  & 32.32   & 25.54 &  11.93  & 7.57  & 5.74\\ 
				{\bf MonoGRNet+LPCG} &  {\bf53.84}   & {\bf37.24}   &  {\bf29.70}  & {\bf16.30}    &   {\bf10.06}   &   {\bf7.86} \\
				 Improvements  & +6.50  & +4.92  & +4.16 &  +4.37 & +2.49 & +2.12 \\ 
				\midrule         
				M3D-RPN \cite{M3D}  &48.56  & 35.94  & 28.59 &  14.53 & 11.07 &8.65 \\ 
				{\bf M3D-RPN+LPCG} & {\bf 62.92}   & \bf{47.14}   & \bf{42.03}  & {\bf 26.17}   &  \bf{19.61}   &  {\bf16.80} \\
				 Improvements  & +14.36  & +11.20  & +13.44 & +11.64 & +8.54 & +8.15 \\ 
				\midrule         
				RTM3D \cite{RTM3D}  & 55.44  &39.24  &33.82 & 13.40 &  10.06 & 9.07 \\ 
				{\bf RTM3D+LPCG} & {\bf65.44}   & {\bf49.40}   &  {\bf43.55}  & {\bf25.23}    &   {\bf19.43}   &   {\bf16.77} \\
				 Improvements  & +10.00  & +10.16  & +9.73 &  +11.83 & +9.37 & +7.70 \\ 
				\midrule         
				RTM3D (ResNet18) \cite{RTM3D}  & 47.78  &33.75  &28.48 & 10.85 &  7.51 & 6.33 \\ 
				{\bf RTM3D (ResNet18)+LPCG} & {\bf 62.98}   & {\bf 45.86}   &  {\bf 41.63}  & {\bf 22.69}    &   {\bf 16.78}   &   {\bf 14.50} \\
				 Improvements  & +15.20  & +12.11  & +13.15 &  +11.84 & +9.27 & +8.17 \\ 
				\midrule         
				GrooMeD-NMS \cite{GrooMeD-NMS}  & 55.62  & 41.07   & 32.89 &  19.67  & 14.32  & 11.27 \\ 
				{\bf GrooMeD-NMS +LPCG} &  {\bf 68.27}   & \bf{50.80}   & \bf{45.14}  & {\bf 27.79}   &  \bf{20.46}   &  {\bf17.75} \ \\
				Improvements  & +12.65  & +9.73  & +12.25 &  +8.12 & +6.14 & +6.48 \\ 
				\midrule
				MonoFlex \cite{MonoFlex}$^*$  & 56.73  & 42.97   & 37.34 &  20.02  & 15.19  & 12.95 \\ 
				{\bf MonoFlex +LPCG} & {\bf 69.16}   & \bf{54.27}   & \bf{48.37}  & {\bf 31.15}   &  \bf{23.42}   &  {\bf20.60} \ \\
				 Improvements  & +12.43  & +11.30  & +11.03 &  +11.13 & +8.23 & +7.65 \\ 
				\bottomrule 
			\end{tabular}
		\end{center}
	\end{table*}

\subsection{Ablation Studies}{\label{sec:abla}}		
	We conduct the ablation studies on KITTI \textit{val} set.
	Because of the space limitation, we provide extra ablation studies in the supplementary material.

~\\
\noindent
{\bf{Different Monocular Detectors.}}
	We plug LPCG into different monocular 3D detectors \cite{MonoGRNet,M3D,RTM3D,GrooMeD-NMS,MonoFlex}, to show its extension ability.
	Table \ref{tab:extension_3d} shows the results. 
	We can see that LPCG obviously and consistently boosts original performances, \textit{e.g.}, $7.57 \rightarrow 10.06$ for MonoGRNet \cite{MonoGRNet}, $10.06 \rightarrow 19.43$ for RTM3D \cite{RTM3D}, and $14.32 \rightarrow 20.46$ for GrooMeD-NMS \cite{GrooMeD-NMS} under the moderate setting (AP$_{3D}$ (IoU=0.7)).  
	Furthermore, we explore the feasibility of using a rather simple model when large data is available.
	We perform this experiment on RTM3D \cite{RTM3D} with ResNet18 \cite{ResNet} backbone, which achieves 46.7 FPS on a NVIDIA 1080Ti GPU. \footnote[1]{\href{https://github.com/Banconxuan/RTM3D}{From RTM3D official implementation.}}
	To the best of our knowledge, it is the simplest and fastest model for monocular 3D detection.
	With employing LPCG, this simple model obtains very significant improvements.
	LPCG endows it (46.7 FPS) with the comparable accuracy to other state-of-the-art models (\textit{e.g.}, GrooMeD-NMS (8.3 FPS)).
	These results prove that LPCG is robust to the choice of monocular 3D detectors.

~\\
\noindent
{\bf{The Number of Annotations.}}{\label{sec:num_ann}}
	We also investigate the impact of the number of annotations.
	We report the results in Table \ref{tab:num_anno}.
	The results indicate that a small number of annotations in LPCG can also lead to high accuracy for monocular 3D detectors.
	For example, the detector using 1000 annotations performs close to the full one (24.89/19.29 \textit{vs.} 25.20/19.61 under the moderate setting).

\section{Conclusion}
	In this paper, we first analyze the label requirements for monocular 3D detection.
	Experiments show that disturbed labels and perfect labels can lead to very close performance for monocular 3D detectors.
	With further exploration, we empirically find that the 3D location is the most important part of 3D box labels.
	Additionally, a self-driving system can produce massive unlabeled LiDAR point clouds, which have precise 3D measurements.
	Therefore, we propose a framework (LCPG), to generate pseudo 3D box labels on unlabeled LiDAR point clouds, to enlarge the training set of monocular 3D detectors.
	Extensive experiments on various datasets validate the effectiveness of LCPG.
	Furthermore, the main limitation of LCPG is more training time due to the increased training samples.

\section*{Acknowledgments}
This work was supported in part by The National Key Research and Development Program of China (Grant Nos: 2018AAA0101400), in part by The National Nature Science Foundation of China (Grant Nos: 62036009, U1909203, 61936006, 61973271), in part by Innovation Capability Support Program of Shaanxi (Program No. 2021TD-05).

%
%
\bibliographystyle{splncs04}
\bibliography{egbib}

\clearpage
\appendix
\begin{center}
	\textbf{\Large Appendix}
\end{center}

 \renewcommand\thesection{\Alph{section}}
 
Due to the space limitation, some details and experimental results are included in this supplementary material as below:
\begin{itemize}
	\item Section {\ref{sec:more_abla} }: More details and ablations.
	\begin{itemize}
  \item[$\bullet$] Section {\ref{sec:details} }: Details of disturbed labels.
  \item[$\bullet$] Section {\ref{sec:LiDAR_det} }: Ablation on different LiDAR detectors.
  \item[$\bullet$] Section {\ref{sec:num_pseudo} }: Ablation on different number of pseudo labels.
\end{itemize}

\item Section {\ref{sec:more_qua} }: Qualitative results.
	\begin{itemize}
  \item[$\bullet$] Section {\ref{sec:qua_res} }: Qualitative results of monocular model detections.
  \item[$\bullet$] Section {\ref{sec:qua_pse} }: Qualitative results of pseudo labels.
\end{itemize}
\end{itemize}

\section{More Details and Ablations}{\label{sec:more_abla}}
\subsection{Details of Disturbed Labels}{\label{sec:details}}
	The labels in Figure 1 in the main text are perturbed by randomly shifting the original value within the percentage range.
	For example, concerning object's 3D location $loc$, $loc^\prime=loc(1+uniform(-\frac{p}{2}, \frac{p}{2}))$, where $uniform(-\frac{p}{2}, \frac{p}{2})$ refers to randomly select a sample from the uniform distribution $\mathcal{U}\left[-\frac{p}{2}, \frac{p}{2}\right]$ ($p$ is 5\%, 10\%, 20\%, 40\% in Figure 1) and $loc^\prime$ is the perturbed value.
	
\subsection{Ablation on Different LiDAR Detectors}{\label{sec:LiDAR_det}}
	This section investigates the effect of different LiDAR detectors in LPCG in the high accuracy mode.
	We employ different LiDAR 3D detectors \cite{PVRCNN,PP,Second,PART}, which are trained by the labeled data.
	They then generate pseudo labels on unlabeled data.
	As shown in Table \ref{tab:different_lidar}, we can see that all LiDAR-based methods bring significant improvements for the monocular detector, and the resulting accuracy is close.
	It indicates that the monocular method is not sensitive to specific LiDAR 3D detectors.

				\begin{table*}
		\begin{center}
		\scriptsize
			\begin{tabular}{l|p{2.0cm}<{\centering}p{2.0cm}<{\centering}p{2.0cm}<{\centering}}
				\toprule  
				\multirow{2}{*}{Monocular Learning Paradigm in LPCG} &   \multicolumn{3}{c}{AP$_{BEV}$ /AP$_{3D}$ (IoU=0.7)$|\scriptstyle R_{40}$} \\ 
				~ &  Easy & Moderate & Hard\\ 
				\midrule         
				M3D-RPN \cite{M3D}  &  20.85/14.53 & 15.62/11.07 &11.88/8.65 \\ 
				\midrule
				PointPillars \cite{PP} + M3D-RPN   & 34.46/26.85  & {\bf 26.74/20.12}  & {\bf 23.75}/17.46 \\ 
				Second \cite{Second} + M3D-RPN  &   {\bf 34.71/28.04}  & 26.07/20.01  & 22.85/17.80 \\ 
				Part-$A^2$ \cite{PART} + M3D-RPN   &   33.28/26.20  & 25.51/20.09  & 22.49/{\bf 17.81}\\ 
				PV-RCNN \cite{PVRCNN} + M3D-RPN  &  33.94/26.17 & 25.20/19.61   & 22.06/16.80 \\
				\bottomrule 
			\end{tabular}
		\end{center}
		\caption{
		Influences of different LiDAR detectors.
		We use different LiDAR 3D detectors to generate pseudo labels.
		The results are close, meaning that LPCG is not sensitive to specific LiDAR detectors.
		}
		\label{tab:different_lidar}
	\end{table*}

\subsection{Ablation on The Number of Pseudo Labels}{\label{sec:num_pseudo}}
	In Table \ref{tab:samples}, we train the monocular 3D detector with different numbers of pseudo labels. 
	We can observe that the performance increases dramatically with more pseudo labels, which means that collecting more unlabeled LiDAR point clouds can further push the performance of monocular 3D detection. 
	Actually, this offline collecting process can be easily achieved in a real-world self-driving system.

	  		  		\begin{table*}
		\begin{center}
		\scriptsize
			\begin{tabular}{l|c|p{2.0cm}<{\centering}p{2.0cm}<{\centering}p{2.0cm}<{\centering}}
				\toprule  
				\multirow{2}{*}{Apporoaches}&\multirow{2}{*}{Samples} &  \multicolumn{3}{c}{AP$_{BEV}$/AP$_{3D}|\scriptstyle R_{40}$}  \\ 
				~&~ &  Easy & Moderate & Hard \\ 
				\midrule 
				\multicolumn{5}{c}{{\textit{\textbf{Under IoU criterion 0.5}}}} \\
				\midrule
				\multirow{7}{*}{M3D-RPN + LPCG} & 100  & 11.18/7.32  & 7.70/4.73   & 6.37/4.06 \\ 
				~&200  & 21.71/17.14  & 16.13/12.32   & 14.26/10.69 \\ 
				~&500  & 31.82/26.82  & 23.31/19.82   & 20.99/17.08\\ 
				~&1000 & 35.09/29.41  & 28.33/23.72   & 25.23/20.93 \\ 
				~&3000 & 54.86/49.75  & 40.11/36.58   & 35.45/32.29 \\ 
				~&10000 & 64.57/59.01  & 47.48/43.97   & 43.56/39.04 \\ 
				~&26057   & {\bf 67.20}/{\bf 62.92}   & \bf{50.52/47.14}   & \bf{46.31/42.03}  \\
				\midrule
				\multicolumn{5}{c}{{\textit{\textbf{Under IoU criterion 0.7}}}} \\
				\midrule
				\multirow{7}{*}{M3D-RPN + LPCG} & 100 & 1.29/0.35  & 0.74/0.21  & 0.62/0.19\\ 
				~&200  & 3.83/1.84  & 3.01/1.20  & 2.28/1.14\\ 
				~&500  &  9.14/4.78  & 6.40/3.35  & 5.52/2.96\\ 
				~&1000 & 11.71/7.48  & 9.23/5.78  & 8.07/4.89\\ 
				~&3000 & 20.96/14.83  & 15.76/10.80  & 13.78/9.76 \\ 
				~&10000 &  31.51/25.05  & 22.76/17.87  & 19.82/15.30  \\ 
				~&26057  & {\bf 33.94}/{\bf 26.17}   &  \bf{25.20/19.61}   &  {\bf22.06/16.80}  \\
				\bottomrule 
			\end{tabular}
		\end{center}
		\caption{
		Influences of the number of pseudo labels. 
		``Samples" in the table denotes the number of training samples, which are generated by pseudo labels.
		All the methods are evaluated with metric $AP|_{R_{40}}$.}
		\label{tab:samples}
	\end{table*}

\section{Qualitative Results}{\label{sec:more_qua}}
\subsection{Qualitative Results of Monocular Model Detections}
\begin{figure*}[h]
		\begin{center}
		\includegraphics[width=1\linewidth]{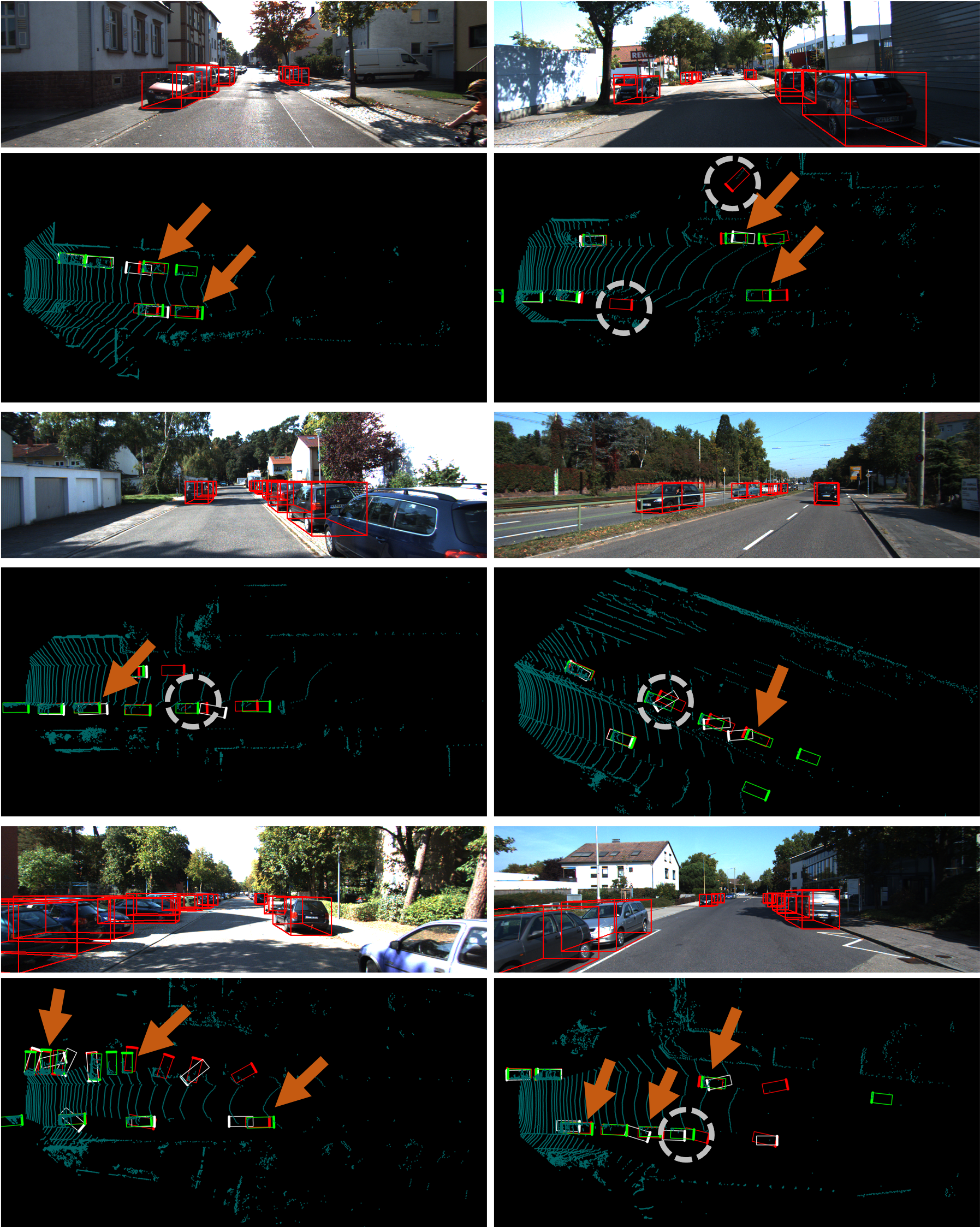}
		\end{center}
		\caption{Qualitative results of M3D-RPN \protect\cite{M3D} trained by our framework (LPCG). {\bf{Green}}: ground-truths. {\bf{Red}}: our predictions. {\bf{White}}: original predictions from M3D-RPN. The bolded side of the box in the bird's-eye-view map refers to the orientation. 
		We can see that our predictions are much more accurate. 
		Note that some predictions are overlapped by ground-truths. 
		We also show failure cases, which are contained in the gray dotted circle.
		Best viewed in color with zoom in.
		}     
		\label{fig: vis}
	\end{figure*}

{\label{sec:qua_res}}
	We provide qualitative results in Figure \ref{fig: vis}.
	We compare the predictions from the original model \cite{M3D} with the ones from the model employing our framework (LPCG). 
	It can be easily seen that our predictions are much more accurate, especially for 3D locations.
	We also show the failure cases, which are usually heavily occluded or faraway objects.
	These objects are hard to be precisely recovered due to the ill-posed nature of monocular imagery.

\subsection{Qualitative Results of Pseudo Labels}
\begin{figure*}[h]
		\begin{center}
								\includegraphics[width=0.94\linewidth]{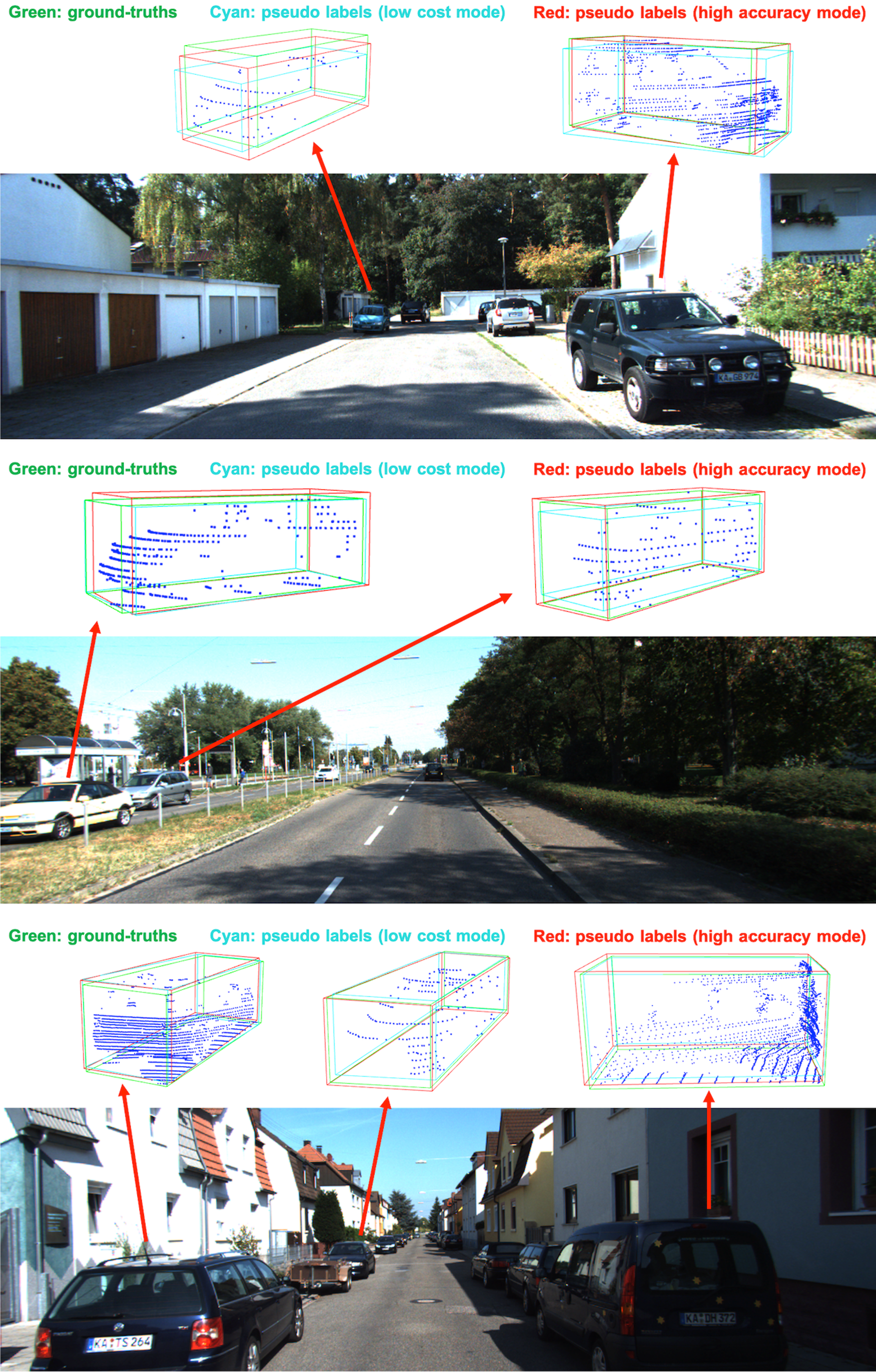}
		\end{center}
		\caption{
		Qualitative comparisons on different labels.
		Thanks to the accurate LiDAR 3D measurements, all types of labels are close, especially for 3D locations.
				}     
		\label{fig: vis_label}
	\end{figure*}

\begin{figure*}[h]
		\begin{center}
		\includegraphics[width=0.865\linewidth]{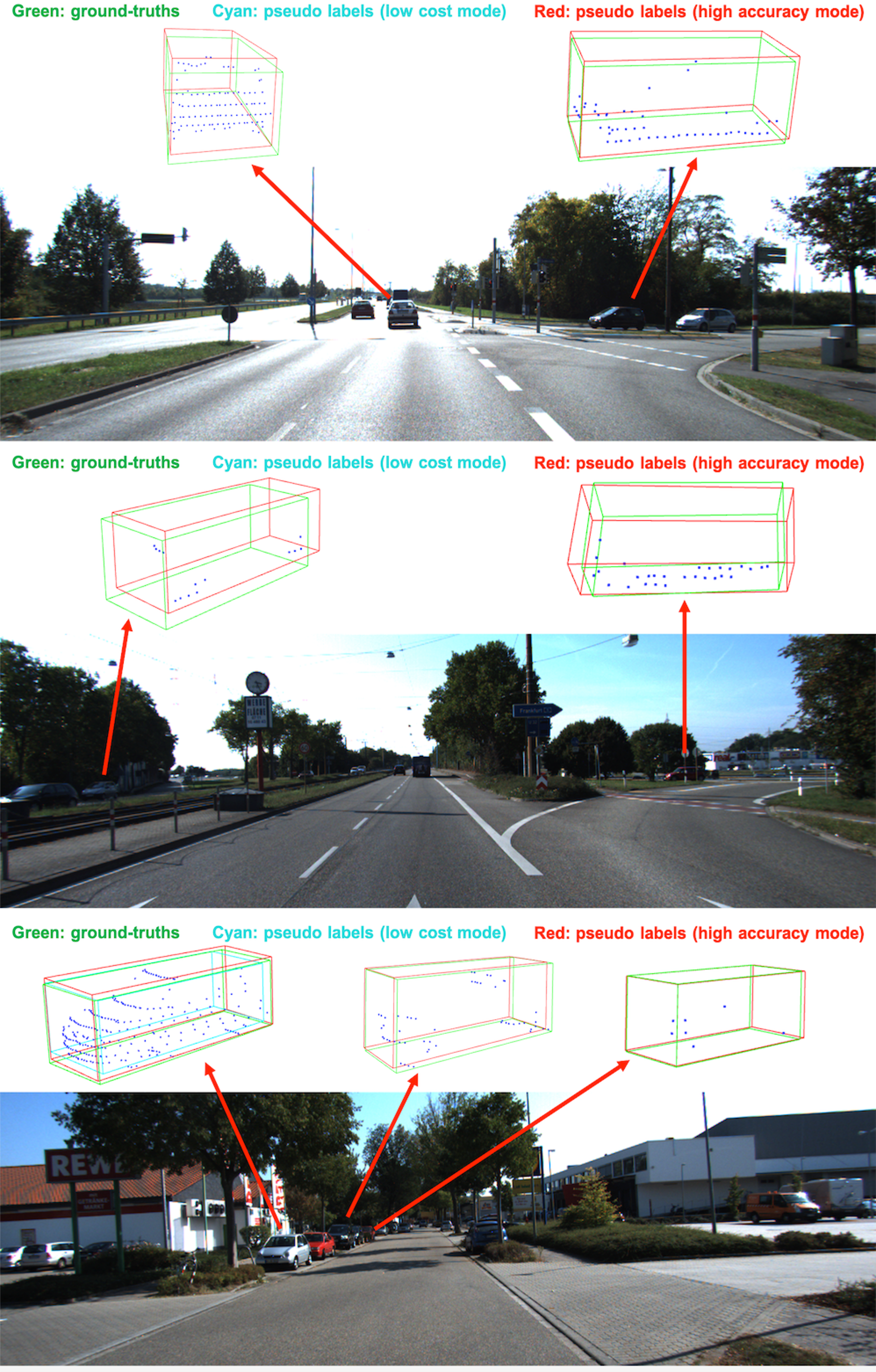}
		\end{center}
		\caption{
		Failure cases of pseudo labels on the low cost mode.
		When the RoI LiDAR point clouds cannot fully describe the object's 3D outline, the geometry-based method cannot recover good 3D box pseudo labels, which are filtered by the dimension constraints.
		Thus many real objects are missed, especially for occluded and faraway objects.
		By contrast, pseudo labels from high accuracy mode still achieve good results because the LiDAR-based network is well-trained, which has studied the latent object pattern. 
				}     
		\label{fig: vis_fail_label}
	\end{figure*}

{\label{sec:qua_pse}}
	To intuitively understand the gap between pseudo labels and manually annotated labels, we illustrate some examples in Figure \ref{fig: vis_label}.
	We can observe that both types of pseudo labels (produced by high accuracy and low cost mode) are close to the manual annotations.
	This accuracy can be attributed to the highly precise 3D measurements of LiDAR point clouds, which give explicit information for obtaining objects' 3D locations.
	However, regarding pseudo labels in the low cost mode, they are produced by the geometry-based method, \textit{i.e.}, finding the minimum bounding box of RoI LiDAR points and then filtering invalid boxes (via the dimension prior).
	When the LiDAR point clouds cannot describe the 3D outline of the object, this geometry-based method will fail since the minimum bounding box is filtered by the dimension constraint.
	Therefore, many real objects are missed.
	These objects usually have few LiDAR points or only have one surface that is captured by the LiDAR device. 
	We show these failure cases in Figure \ref{fig: vis_fail_label}.

\end{document}